\documentclass{article}

\usepackage[utf8]{inputenc} 
\usepackage[T1]{fontenc}    
\usepackage{hyperref}       
\usepackage{url}            
\usepackage{booktabs}       
\usepackage{amsfonts}       
\usepackage{nicefrac}       
\usepackage{microtype}      
\usepackage{amsmath}
\usepackage{bbm}
\usepackage{tikz}
\usepackage{amsthm}
\usepackage{graphicx}
\usepackage{epstopdf}
\usepackage{authblk}

\usepackage{geometry}
\providecommand{\theoremname}{Theorem}
\providecommand{\propositionname}{Proposition}
\newtheorem{thm}{\protect\theoremname}
  \newtheorem{prop}[thm]{\protect\propositionname}

\title{\bf Bayesian Transfer Reinforcement Learning with Prior Knowledge Rules }

\author[1]{Michalis K. Titsias}
\author[1]{Sotirios Nikoloutsopoulos}
\affil[1]{Athens University of Economics and Business}

\date{}

\begin{document}

\maketitle

\begin{abstract}
We propose a probabilistic framework to directly insert prior knowledge in reinforcement learning (RL) algorithms  
by defining the behaviour policy as a Bayesian posterior distribution. Such a posterior 
combines task-specific information with prior knowledge, thus allowing to achieve transfer 
learning across tasks.  The resulting method is flexible and it can be easily incorporated 
to any standard off-policy and on-policy algorithms, such as those based on temporal differences and policy gradients. 
We develop a specific instance of this Bayesian transfer RL framework by expressing prior knowledge as general 
deterministic rules that can be useful in a large variety of tasks, such as navigation tasks. 
Also, we elaborate more on recent probabilistic and entropy-regularised RL by developing a 
novel temporal learning  algorithm and show how to combine it with Bayesian transfer RL. Finally, we demonstrate 
our method  for solving mazes and show that significant speed ups can be obtained.      
 \end{abstract}

\section{Introduction}

Humans use prior knowledge  to act  and learn  from  sequential decision making tasks.   
Prior knowledge is important since it allows for efficient solving new tasks with minimal additional learning 
and computational effort.    
Such knowledge is gradually  built through interaction with the world through a process that can be credited to the  ability of the human brain  to  
 understand about how the world works and efficiently organise such knowledge for future use.  
A simple, but highly interpretable,  way to describe  knowledge is by using 
a set of rules that have been induced from past experience. These rules can be used to guide human decision 
making in new tasks by essentially shaping future behaviour so that sample efficient learning is achieved. 
Such rules could be inter-related, e.g.\ hierarchically ordered, and probabilistic. 
For instance, from early childhood someone can pick up the rule  "things fall" and also 
"certain things made of glass when they fall they can break".   By knowing already that glass can break, a child playing with a toy made of glass
will probably avoid actions that can result in breaking the toy, until  
the game becomes less interesting and breaking or not breaking the toy are not that different in terms of enjoying the game.  
The main aspect  of this and many similar examples is that  past knowledge is used (together with task-specific information such as 
reward) to shape current behaviour.

Motivated by the above we propose a probabilistic framework, referred to as  Bayesian transfer reinforcement learning (RL), to directly 
insert prior knowledge in RL algorithms  by defining the behaviour policy as a Bayesian posterior distribution.
 Such a posterior combines task-specific information with prior knowledge, thus allowing to achieve transfer 
learning across tasks. In other words, we perform transfer or meta-learning RL \cite{Duan16, Wang16, Stadie17, Gupta18, Teh17}
through a Bayesian behaviour policy while other aspects of the  RL learning algorithm, such as whether  we  
are based on value function estimation or direct policy estimation, can remain largely unchanged. 
Further,  the incorporation of prior knowledge can be arbitrarily complex in the sense that the behaviour policy can be constructed 
using high-order non-Markovian relationships associated with the history of observed states and actions, memory,  deterministic rules  etc.  
In the current  implementation we are based on simple deterministic rules that are incorporated into a given task using 
a procedure called prior realisation,  while more advanced methods for more drastically learning such rules is left for future research. 

 Furthermore,  we illustrate our method using a probabilistic formulation of RL \cite{Todorov2009,Kappen2012,Toussaint:2009, azar11a, Rawlik:2013}  
 since it can lead to algorithms, such as  entropy-regularised methods \cite{WilliamsPeng91, Peters:2010, Schulman:2015, mniha16}, that can naturally capture  uncertainty.
  For probabilistic RL we also develop a novel temporal learning algorithm, similar to those in  \cite{haarnoja17a, Fox:2016, asadi17a}, that for the tabular case approximates 
  a set of linear equations by sequentially  updating a certain state-action value function.  In the experiments we apply this temporal learning algorithm jointly with 
 Bayesian transfer RL for solving mazes and show that significant speed ups can be obtained.  




\section{Bayesian transfer reinforcement learning}
   

Consider a RL problem  where at
each time step $t$ an agent performs an action $\alpha_t \in \mathcal{A}$, observes a new state $s_{t+1} \in \mathcal{S}$ drawn from $p(s_{t+1} | s_t, \alpha_t)$
and receives reward $r_t = r(s_t,\alpha_t)$. The transition density  $p(s_{t+1} | s_t, \alpha_t)$  and the reward function $r(s_t,\alpha_t)$ can be unknown or partially known to the agent. 
By taking actions and observing states the agent wishes to learn a policy that maximises  a measure that depends on future reward, such 
as expected discounted reward or average reward;  see \cite{Sutton:1998} for full details.  The policy according to which actions are taken is
the behaviour policy. In off-policy RL the behaviour policy is different from the actual policy that the agent eventually learns, while in  on-policy  
RL the two policies coincide.  Next  we introduce a probabilistic Bayesian procedure that introduces prior knowledge 
into the structure of the behaviour policy. We discuss separately the two distinct cases, i.e.\  off-policy and on-policy learning.      

Off-policy RL is usually based on temporal difference methods, such as $Q$-learning \cite{Watkins:89}. Let us assume 
that the action space $\mathcal{A}$ is discrete and the agent has an estimate of the state-action value function $Q(s_t, \alpha_t)$  and it can update such estimate
 at each iteration. 
A behaviour policy based on the current estimate of the $Q$ function typically takes the 
softmax or Boltzmann form, 
$$
\pi (\alpha_t | s_t )  =  \frac{\exp\{  \beta Q(s_t, \alpha_t) \}}
{ \sum_{\alpha \in \mathcal{A}}  \exp\{  \beta Q(s_t, \alpha)  \}},
$$
 where $\beta \geq 0$ is a hyperparameter that specifies  the amount of uncertainty; when $\beta=0$ the above becomes the uniform 
 distribution while for large $\beta \gg 0$ actions will be chosen based on the maximum $Q$ value.  The above policy  uses only 
 task-specific information, i.e.\ the state-action value function for the task at hand. To combine this with prior knowledge we can  express the following Bayesian 
 behaviour policy, 
 \begin{align}
 q( \alpha_t | s_t, \mathcal{M}_t) & \propto \underbrace{   exp\{  \beta Q(s_t, \alpha_t) \} }_{\text{task-specific information} }  \times 
 \underbrace{ f(\alpha_t; \mathcal{M}_t) }_{ \text{prior knowledge} }.
 \label{eq:bahaviourPolicy}
 \end{align}
 Here, the non-negative function $f(\alpha_t; \mathcal{M}_t) \geq 0$ is an unnormalised probability distribution 
 that  takes high values  for actions $\alpha_t$ that  conform well with  the current prior knowledge and small values otherwise.   
 $\mathcal{M}_t$  represents all stored information needed to express prior knowledge and compute values of $f(\cdot)$. 
 For instance, $\mathcal{M}_t$ can include tunable parameters, histories of previous states and actions, reward values  etc. 
 In section \ref{sec:deterministic} we give an example of how to construct  $f(\alpha_t; \mathcal{M}_t)$ using indicator functions 
 and simple deterministic rules.  The $\beta$ hyperparameter allows  to balance between task-specific information and  prior knowledge.  
  
The above Bayesian behaviour policy offers a prior-informed exploration that can be used for off-policy RL, i.e.\   to learn the state-action value function. 
However, unlike standard exploration mechanisms such e-greedy procedures the policy in \eqref{eq:bahaviourPolicy} may completely 
ignore uninteresting parts of the state space, i.e.\ the ones unfavored by the prior $f(\alpha_t; \mathcal{M}_t)$. Clearly, this can be highly 
desirable (given that these uninteresting parts truly do not contain information about the optimal policy) and arguably it is the only way to achieve sample efficient learning in very large or continuous state spaces. 

A second  way to use Bayesian behaviour policies is as part of on-policy algorithms. Such schemes 
are typically used with direct policy optimisation based on algorithms such as {\sc REINFORCE} \cite{Williams:1992} and actor-critic methods. To  extend our method to cover
such cases  we can modify the policy  from \eqref{eq:bahaviourPolicy} according to 
\begin{align}
 q( \alpha_t | s_t, \mathcal{M}_t) & \propto  p(\alpha_t | s_t)   \times  f(\alpha_t; \mathcal{M}_t) .
 \label{eq:bahaviourPolicy2}
\end{align}
 where  $p(\alpha_t | s_t) $  is a policy that aims at capturing task-specific information while  $ f(\alpha_t; \mathcal{M}_t)$ is the  same prior 
 that appears in \eqref{eq:bahaviourPolicy}. $q( \alpha_t | s_t, \mathcal{M}_t )$ consists of the overall policy 
 that can be optimised with respect to the parameters of $p(\alpha_t | s_t)$ and possibly of  any parameters of the prior $f(\alpha_t; \mathcal{M}_t)$.

 
 In the next section we  present a simple example of how to construct  $f(\alpha_t; \mathcal{M}_t)$ using indicator functions 
 and simple deterministic rules.

 \subsection{Incorporate prior knowledge using deterministic rules  \label{sec:deterministic}}

Prior knowledge in RL means that we know something about how the world works. Specifically, this translates to 
knowing something about the environmental transition densities $p(s_{t+1}|s_t, \alpha_t)$  and possibly 
the reward function $r(s_t, \alpha_t)$.  We shall focus on 
general purpose prior knowledge, phrased as simple intuitive rules, 
that can be useful to a large number of tasks. 
An example of general purpose prior knowledge is that of 
knowing that the environmental transition densities $p(s_{t+1}|s_t, \alpha_t)$ are 
deterministic and/or stationary, which can already be a truly powerful 
prior information that can lead to practical  knowledge rules as the following one:  

\vspace{2mm}

\emph{In a roughly deterministic and stationary world, past plans that resulted in no progress for solving a task 
need to be tried out less frequently in the future}  

\vspace{2mm}

Humans possibly pick up such rule from past experience and particularly by observing 
that the world is largely deterministic and the rules about how the world works
do not unpredictably change.\footnote{I.e.\ the world at the time scale where 
one solves a certain task is often stationary.}  
To see an example of how a human applies this rule, suppose a car driver tries to 
reach a certain destination. In case the driver starts at state/location 
$s_0$, follows a certain route (consisted of several locations and actions) and 
returns to the same location $s_0$, he knows that trying 
again the same route is largely pointless. In the remaining of this section we present a way to 
implement simple versions of the above rule and incorporate them into the prior $f(\alpha_t; \mathcal{M}_t)$.  

Given that a plan has certain length, corresponding to the number of actions comprising the plan,
we can define M-order rules that conform with the above general rule and where $M$ corresponds to length. 
We shall focus on the following 1-order and 2-order rules since they are the simplest ones:

\begin{itemize}

\item 1-order rule: Suppose we are at state $s_t=s$, apply action $\alpha_t=\alpha$  and the state remains 
unchanged, i.e.\  the next state is $s_{t+1}=s$.  Then the plan of taking action $\alpha$ whenever we are at state 
$s$ should never be tried, unless there is reward for staying at $s$. 

\item 2-order rule: Suppose we are at state $s_t=s$, apply action $\alpha_t=\alpha$, move to the state $s_{t+1}=s' \neq s$ 
and by applying a second action $\alpha_{t+1} = \alpha'$ we return to the initial state $s_{t+2} = s$.
 Then the plan of taking the sequences of actions $(\alpha, \alpha')$ whenever we are at state 
$s$ should never be tried, unless there is reward for staying at $s$ or $s'$.

\end{itemize}       
  
Intuitively, the first case above describes situations like \emph{trying to walk through a wall} or \emph{a fly trying to go through a window}. 
The second case  describes \emph{pairs of undoing or opposite actions}  such as (left,right) and (up,down). 
An undoing pair is universal when this holds for any state $s$, which is often the case for many tasks such as navigation tasks. 
Notice, that two actions in order to really be undoing pairs they must result in returning to the same state after 
consecutively applying both actions.
 For instance, the (left,right) pair in a navigation task, such as escaping 
from a maze, is truly an undoing pair,  but this might not hold for other tasks such as when playing an  Atari game (where the state of the game such as 
the location of an object might not be the same after applying a left and then a right move) or in continuous control problems.  
For these latter cases the above rules might hold in an approximate or probabilistic manner; see Section \ref{sec:discussion} for further discussion.  

We can incorporate the above 1-order and 2-order rules in the prior $f(\alpha_t; \mathcal{M}_t)$ using indicators functions  (essentially simple if-then-else rules).  
For any given task,  we need to store in memory $\mathcal{M}_t$ all cases where the 1-order rule applies and all pairs of undoing actions (assuming that such pairs 
are universal).  Notice that regarding the latter case we can express all indicator functions as a binary matrix  $g(\alpha_t; \alpha_{t-1})$ that takes the value zero in any entry
where the current action $\alpha_t$ and the previous action $\alpha_{t-1}$ are undoing pairs.  
The process of gradually building this memory $\mathcal{M}_t$ with all these cases is referred to as \emph{prior realisation} 
(since it realises our prior knowledge to the specifics of a given task)  and it is carried out through actual experience,  i.e.\ as the agent interacts with 
the environment.    
 
\section{Probabilistic reinforcement learning  \label{sec:probabilisticRL} }
 
Bayesian transfer RL could work in conjunction  with standard off-policy and on-policy algorithms.  
However, it can more naturally be used together with algorithms that represent uncertainty when estimate 
task-specific policies, such soft-$Q$ learning  \cite{haarnoja17a} and related methods  \cite{Fox:2016, asadi17a, azar11a, Rawlik:2013} as well as policy gradient methods 
with entropy regularization \cite{WilliamsPeng91, Peters:2010, Schulman:2015, mniha16}.
Therefore here we re-visit the probabilistic RL framework  \cite{Todorov2009,Kappen2012,Toussaint:2009, azar11a, Rawlik:2013, Levine18}  
and introduce also a novel temporal learning 
algorithm. In the experiments we apply this algorithm together with Bayesian transfer RL for solving mazes. 
   
Consider an episodic RL setting, where we start at state $s_0$ and we generate a sequence of states and actions  
according to the joint distribution 
\begin{equation} 
p(\alpha_{0:h-1}, s_{1:h} | s_0 ) =  \prod_{t=0}^{h-1} \pi_{0} (\alpha_t | s_t) p(s_{t+1}|s_{t}, \alpha_t).   
\label{eq:prior}
\end{equation}
The episode ends when we reach a terminal state
$s \in \mathcal{T} \subset \mathcal{S}$.  $\pi_{0} (\alpha_t | s_t)$ is  a
baseline stochastic policy which could be a very broad distribution, e.g.\ for  discrete action spaces it could be 
uniform. We introduce rewards $r_t = r(s_t,\alpha_t)$ associated with a certain task and assume that 
$r_t$ is a deterministic function of the state-action pair $(s_t,\alpha_t)$ (extending to random rewards 
where $r_t \sim p(r_t | s_t, \alpha_t)$ is straightforward).  
The rewards aim at constraining the above joint distribution 
towards state-action 
sequences leading to high values of accumulated reward $\sum_{t=0}^{h-1} r_t$. 
To incorporate such constraint  into the joint distribution
we introduce the exponentiated reward factors  $\exp( \beta r_t )$, where $\beta>0$ is a hyperparameter, and consider the 
factorisation
\begin{align} 
f(r_{0:h-1}, \alpha_{0:h-1}, s_{1:h} | s_0)  & = \prod_{t=0}^{h-1}
\exp\left( \beta r_t \right)
 \pi_{0} (\alpha_t | s_t) p(s_{t+1}|s_{t}, \alpha_t). 
\label{eq:factorization}
\end{align}  
This factorization does not define a joint probability distribution since the factors 
$\exp\left( \beta r_t \right)$ are not distributions but soft constraints that favour  high reward values. 
Several authors \cite{Rawlik:2013, Levine18} interpret 
each term $\exp( \beta r_t )$ as the probability $p(\mathcal{O}_t=1 |s_t, \alpha_t)$ of an auxiliary binary variable  
$\mathcal{O}_t$ so that  $\exp( \beta \sum_{t=0}^{h-1} r_t )$ is precisely the likelihood $\prod_t p(\mathcal{O}_t=1 |s_t, \alpha_t)$ of all these binary 
variables taking the value one. However, this interpretation is rather artificial and also restrictive since it is 
valid only when $r_t \leq 0$ so that $\exp( \beta r_t ) \in [0,1]$. Instead,  here we assume that $r_t$ takes arbitrary 
finite values and view the overall factorization in \eqref{eq:factorization} as a potential function (similarly to 
undirected graphical models) that allows us to define the following  posterior distribution, 
\begin{align} 
p(\alpha_{0:h-1}, s_{1:h} | s_0, r_{0:h-1} ) 
=  \frac{1}{\mathcal{Z}}  \exp\left( \beta \sum_{t=0}^{h-1} r_t \right) \left[ \prod_{t=0}^{h-1} \pi_{0} (\alpha_t | s_t) p(s_{t+1}|s_{t}, \alpha_t) 
\right], 
\label{eq:posterior}
\end{align}  
where $\mathcal{Z}$ denotes the normalizing constant. 
The hyperparameter $\beta>0$ determines the relative strength of the rewards factor 
$\exp( \beta \sum_{t=0}^{h-1} r_t )$
versus  $p(\alpha_{0:h-1}, s_{1:h} | s_0)$. 

In order to utilise the posterior distribution  
in \eqref{eq:posterior}
for RL  we need to compute the optimal policy that is consistent with the full posterior. 
%
At any given time all past states, actions and rewards have been observed and the agent needs 
to take the current action by marginalising out all possible future sequences 
that could be possibly realised after taking this action. 
Given that we are at state $s_t$ we are interested in computing the  
marginal posterior distribution over action $\alpha_t$ conditioning on all rewards $r_{0:h-1}$ 
but also on the full history of all past states and actions $(s_{0:t-1}, \alpha_{0:t-1})$,   
\begin{equation}
p(\alpha_t | s_{0:t-1}, \alpha_{0:t-1}, r_{0:h-1}) = p(\alpha_t | s_t, r_{t:h-1}). 
\label{eq:postPolicy}
\end{equation}
The simplification in the r.h.s.\ is because  when conditioning on the current state $s_t$
the action $\alpha_t$ becomes independent from all previous states, actions and rewards, 
and it depends only on the future rewards. Of course,  this simplification is due
to the Markovian nature of the model.   
From probabilistic inference perspective 
$p(\alpha_t | s_t, r_{t:h-1})$ is the optimal policy. Such policy satisfies  a 
Bellman-type of recursive equation as stated next.  

\begin{prop}
For the posterior in \eqref{eq:posterior} the optimal policy  $p(\alpha_t | s_t, r_{0:h-1})$ 
is computed as 
\begin{equation}
p (\alpha_t|s_t, r_{t:h-1}) = \frac{B(s_t,\alpha_t)}{\sum_{\alpha_t} B(s_t,\alpha_t)} = 
\frac{B(s_t,\alpha_t)}{A(s_t)}   
\label{eq:optimalPolicy}
\end{equation}
\begin{equation}
B(s_t,\alpha_t)  =  e^{ \beta r_t } \pi_0 (\alpha_t | s_t)  \sum_{s_{t+1}}  p(s_{t+1}|s_t,\alpha_t)  
A(s_{t+1}), \ \ A(s_{t+1}) =  \sum_{\alpha_{t+1}}  B(s_{t+1},\alpha_{t+1}), \ A(s)=1 \forall s \in \mathcal{T}.  
\nonumber 
\label{eq:Betarecursion}
\end{equation}
\end{prop} 
The proof is given in the Appendix. From the above
we can also conclude that $A(s_t)$ satisfies the recursion 
$A(s_t)  = \sum_{\alpha_t} e^{ \beta r_t } \pi_0 (\alpha_t | s_t)  \sum_{s_{t+1}}  p(s_{t+1}|s_t,\alpha_t)  
A(s_{t+1})$. $B(s_t,\alpha_t)$ can be considered as a state-action value function 
while $A(s_t)$ as a state value function. When the environmental dynamics $p(s_{t+1}|s_t,\alpha_t)$ are known 
(and each $s_t$ and $\alpha_t$ take discrete values) we can compute the $B$ function, and 
subsequently the optimal policy, by unfolding the recursion backwards or by applying a  
linear system solver. However for RL, where $p(s_{t+1}|s_t,\alpha_t)$ are unknown, we will 
need to apply stochastic approximation to learn from actual experience as 
discussed shortly.  
  
The state-action value function $B(s_t,\alpha_t)$ connects with the optimal $Q$ function in the regular 
reinforcement learning as shown in the following statement.  

\begin{prop}
Suppose discrete state and action spaces, deterministic transitions such that $p(s' |s,\alpha) = \delta(s' - d(s,\alpha))$
and $\pi_0 (\alpha | s)>0$ for any $s,\alpha$. Then as $\beta \rightarrow \infty$, $\frac{1}{\beta} \log B(s,\alpha)$ converges 
to the state-action value $Q_*(s,\alpha) = r + \max_{\alpha}\{ Q_*(d(s,\alpha),\alpha)\}$ 
associated with the optimal deterministic policy in regular reinforcement learning.
\end{prop} 
The proof is given in the Appendix. Clearly also for  the deterministic environmental transitions 
$ (1/\beta) \log A(s_t)$ converges to the value function $V(s_t)$. For stochastic environmental transitions 
the state-action function  $B(s_t,\alpha_t)$ will be generally different from $Q(s_t,\alpha_t)$ 
in regular RL. For instance, observe that 
$$
(1/\beta) \log B(s_t,\alpha_t)  =  r_t  +  (1/\beta) \log \pi_0 (\alpha_t | s_t)  
+ (1/\beta) \log \sum_{s_{t+1}}  p(s_{t+1}|s_t,\alpha_t)  A(s_{t+1}),
$$     
where the expectation under $p(s_{t+1}|s_t,\alpha_t)$ is inside the log while in the
$Q$ function is outside.  As discussed in \cite{Levine18} this can result in an optimistic policy 
that is unrealistic in most control problems. To overcome this, we could replace the optimal policy with a variational approximation obtained 
by imposing the actual transition densities $p(s_{t+1}|s_t,\alpha_t)$ as part of the approximation; see 
 \cite{Levine18}  and  \cite{Rawlik:2013}  for full details.

Given that the logarithm of $B(s_t,\alpha_t)$  connects with the  
$Q$ function in regular RL led many authors to derive temporal difference 
stochastic approximation algorithms, such soft $Q$-learning and $G$-learning that operate in the log space 
\cite{haarnoja17a, Fox:2016, asadi17a, azar11a, Rawlik:2013}. 
%
 However, from probabilistic inference viewpoint another direct way to 
apply stochastic approximation is to be based on the initial linear recursions of Proposition 1.  
    
More precisely, for discrete states and actions we wish to directly 
approximate the Bellman equation in Proposition 1 so that to stochastically 
approximate  the state-action value $B(s_t,\alpha_t)$. 
Notice that in this discrete setting 
the whole function $B(s,\alpha)$ reduces to a table 
of size $|\mathcal{S}| \times |\mathcal{A}|$. 
For any terminal state $s \in \mathcal{T}$ we set $B(s,\alpha) = 1/|\mathcal{A}|$ and 
the remaining values to arbitrary strictly positive values. 
Then, at each time step $t$ we perform an action based on some behaviour policy and we
obtain the following stochastic estimate for the entry $B(s_t,\alpha_t)$,
$$
\widetilde{B}(s_t,\alpha_t) =  e^{ \beta r(s_t,\alpha_t) } \pi_0 (\alpha_t | s_t) 
 \sum_{\alpha_{t+1}} B(s_{t+1},\alpha_{s+1})
$$
and then we do a stochastic optimization update\footnote{For numerical stability the update is performed based on the logsumexp trick 
by keeping track of the logarithm of  $B(s_t,\alpha_t)$.}
\begin{equation}
B(s_t,\alpha_t) = (1 - \rho_t) B(s_t,\alpha_t)  + \rho_t \widetilde{B}(s_t,\alpha_t),
\label{eq:tableB} 
\end{equation}
where $\{\rho_t\}$ is the learning rate sequence satisfying the standard Robins-Monroe conditions
and where for each terminal state $s \in \mathcal{T}$ the values are fixed to $B(s,\alpha)=1/ |\mathcal{A}|$ 
which ensures that $\sum_{\alpha} B(s,\alpha) = A(s)=1$. Repeated application of the 
update in \eqref{eq:tableB} stochastically approximates a set of linear equations. 


If we wish to combine the above temporal learning algorithm with Bayesian transfer RL from the previous section 
we simply need to consider an off-policy algorithm where the actual experience of the RL  agent is collected based on the following 
behaviour policy 
$$
 q( \alpha_t | s_t, \mathcal{M}_t)  \propto  B(s_t,\alpha_t)  \times  f(\alpha_t; \mathcal{M}_t),
$$   
where $B(s_t,\alpha_t)$ is the current estimate of the state-action value,  updated according to  \eqref{eq:tableB}, 
and  $f(\alpha_t; \mathcal{M}_t)$ is the prior that allows to transfer past knowledge.   In the next section we consider the prior 
 that represents the deterministic rules from Section \ref{sec:deterministic} and apply the overall scheme for solving mazes. 
    
\section{Experiments} 

\begin{figure*}[!htb]
\centering
\begin{tabular}{ccc}
{\includegraphics[scale=0.14]
{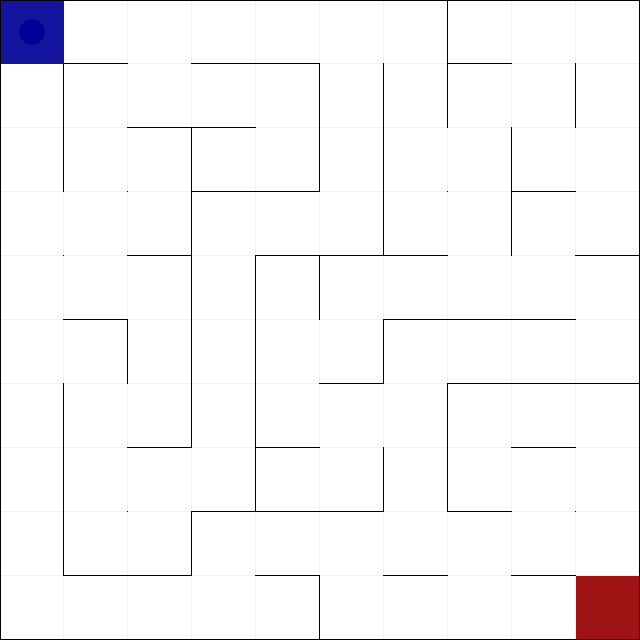}} &
{\includegraphics[scale=0.14]
{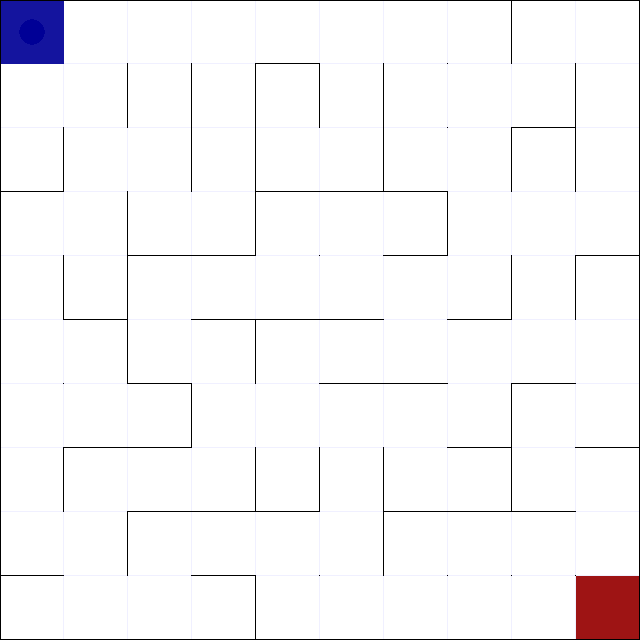}} &
{\includegraphics[scale=0.14]
{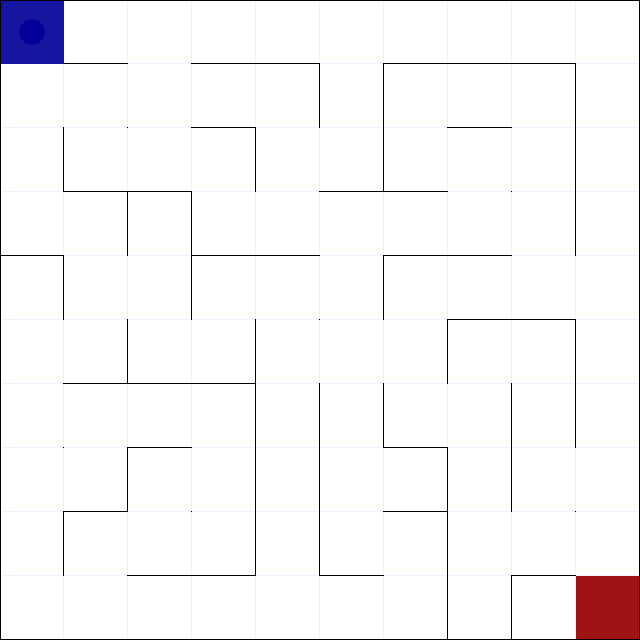}}
{\includegraphics[scale=0.14]
{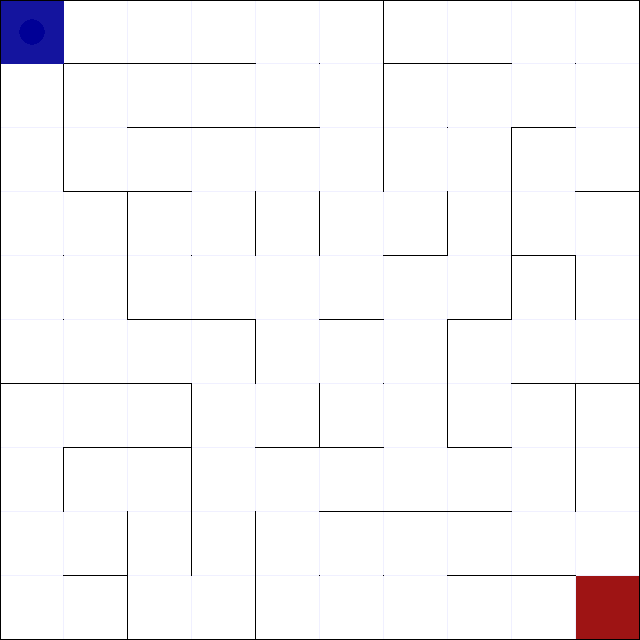}}
\end{tabular}
\caption{Four mazes from the set of 100 random mazes.} 
\label{fig:games}
\end{figure*}

Here, we demonstrate the Bayesian transfer RL algorithm for solving mazes implemented using openai gym. We generated 100 
random mazes which consist of $10 \times 10$ grids, such as those shown in Figure  \ref{fig:games}, where the agent starts at blue top-left corner and wishes to 
reach the red bottom-right corner. 
We assume that at each state there are four possible actions (up,down,right,left), the environmental dynamics are determinist and the semantics of the four actions are the same across 
all tasks.  A reward of $1$ is given when the agent reaches the goal, while for
 every step in the maze the agent recieves a reward of value $-0.001$. For all experiments below 
we fix $\beta=1000$ based on the simple heuristic that a good value  
is such that $\beta \approx O(\frac{1}{|r|})$ where $r$ is a typical reward value.   

We implemented and compared three different methods: (i) The temporal learning algorithm  introduced
in Section \ref{sec:probabilisticRL}  using as behaviour policy $p(\alpha_t | s_t) \propto B(s_t, \alpha_t)$
which corresponds to an on-policy procedure. We refer to this method as {\sc no-prior} since no prior knowledge
is used. (ii) The off-policy scheme where the behaviour policy 
is $ q( \alpha_t | s_t, \mathcal{M}_t)  \propto  B(s_t,\alpha_t)  \times  f(\alpha_t; \mathcal{M}_t)$ and where 
the prior accounts only for the 1-order rule from Section \ref{sec:deterministic}. We refer to this method as 
{\sc 1-prior}. (iii)  A scheme similar to (ii) but where the prior accounts for both the 1-order  and the 2-order  rule from Section  
 \ref{sec:deterministic}. This third method is referred  to as {\sc 1\&2-prior}.  For   the {\sc 1-prior} case the prior realisation process is 
 such that the memory $\mathcal{M}_t$ starts from the empty set and is updated on the fly by inserting state-action pairs $(s,\alpha)$
 that result in no change in the state. Subsequently, the actions in $\mathcal{M}_t$ given that we are in the corresponding state 
 are never taken. For the  {\sc 1\&2-prior} case $\mathcal{M}_t$ also includes the two pairs of opposite actions which are assumed to be universal (see Section \ref{sec:deterministic}) 
 and they are  quickly discovered in the first few moves when solving the first maze, so that this knowledge is transferred through the behaviour policy 
 to all subsequent iterations and different mazes. 
 
\begin{figure*}[!htb]
\centering
\begin{tabular}{c}
{\includegraphics[scale=0.5]
{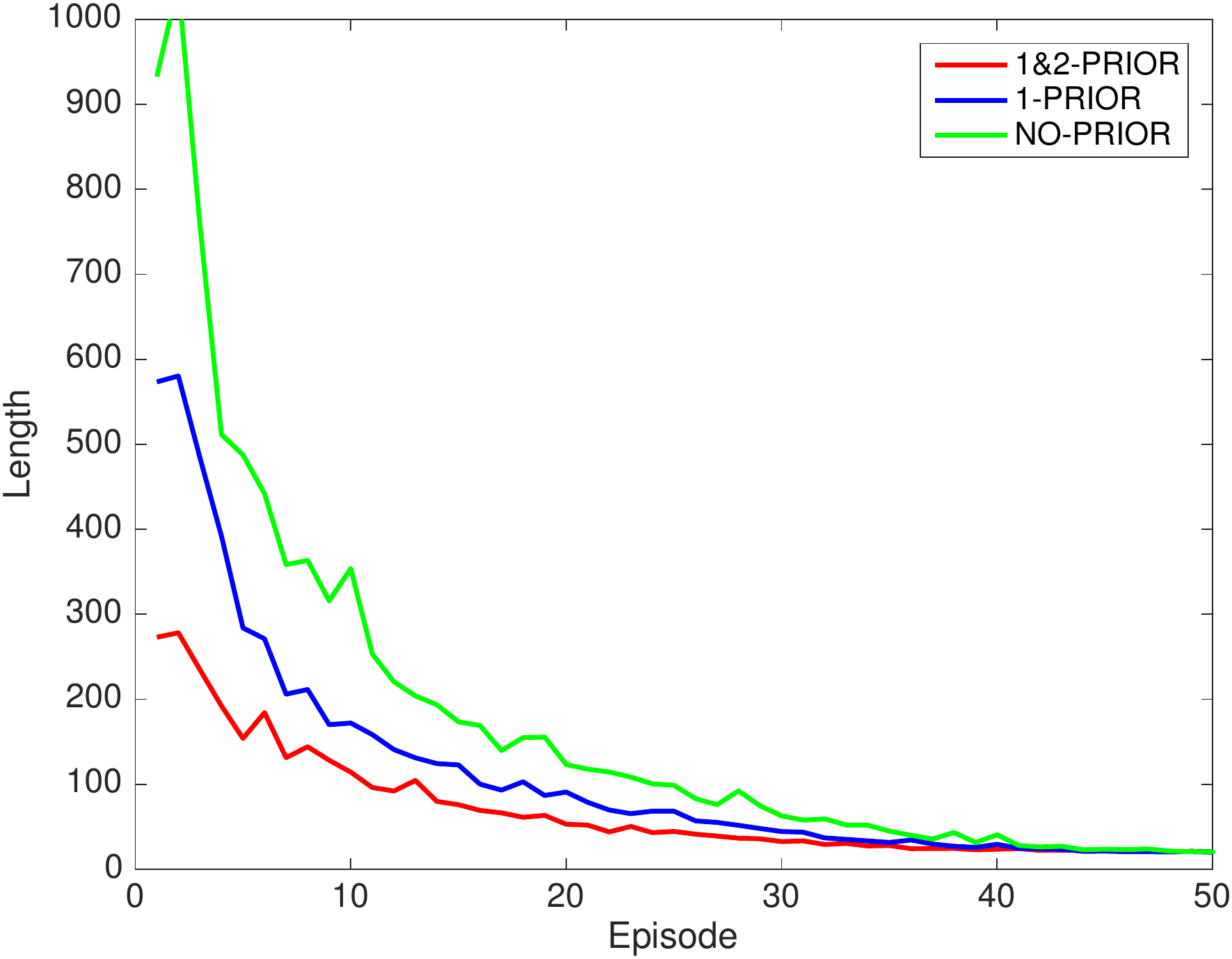}} 
\end{tabular}
\caption{The horizontal axis corresponds to the episode (which completes when a maze is solved) while the vertical 
axis shows the length of the episode, i.e.\ the number of steps the agent took in order to exit the maze. All three curves 
are averages across all 100 random mazes.} 
\label{fig:curves}
\end{figure*}

 Figure \ref{fig:curves} shows average performance for all three methods. Clearly, by adding prior knowledge  
 in the behaviour policy learning is speeded up in a systematic way so that {\sc 1-prior} is better than 
 {\sc no-prior}  and  {\sc 1\&2-prior}  is better than  {\sc 1-prior}.

\section{Discussion \label{sec:discussion}} 

We proposed a framework to carry out transfer learning in RL  
through Bayesian behaviour policies that can combine task-specific information with prior knowledge. 
The resulting method is very general and it can work together 
with  standard off-policy and on-policy RL algorithms, although 
probabilistic versions of such algorithms are the most suitable due to
their natural ability to capture uncertainty.   

We showed how to  represent prior knowledge using intuitive deterministic rules 
and demonstrated this for solving mazes. However, for more realistic applications these rules 
will not hold in a fully determinist manner because of the uncertainty and high complexity of the real-world environments.  
 Thus, one main direction for future work is to define soft or probabilistic relaxations of the initial deterministic rules so that to deal with real-world applications.  
 For instance, given that the constraints associated with undoing pairs of actions can be represented using a binary matrix 
 $g(\alpha_t; \alpha_{t-1})$ (see Section \ref{sec:deterministic}),  by relaxing this and parametrising each entry of this matrix with the sigmoid 
 function we could potentially learn arbitrary transition relationships between consecutive actions. 
 Similarly, it would be useful to investigate whether it is possible to more drastically learn prior rules 
(or the full structure of the prior $f(\alpha_t;\mathcal{M}_t)$) based on gradient-based optimisation 
and deep learning. 


\appendix 
 
\section{Proofs} 

We first prove Proposition 1 by showing how to compute  $B(s_0,\alpha_0)$ while the general case 
is similar. 
Given that we start at state $s_0$ we wish to compute the optimal 
policy $p(\alpha_0 | s_0, r_{0:h-1})$: 
$$
p (\alpha_0|s_0, r_{0:h-1}) = \frac{B(s_0,\alpha_0)}{\sum_{\alpha_0} B(s_0,\alpha_0)} = 
\frac{B(s_0,\alpha_0)}{A(s_0)} 
$$
$B(s_0,\alpha_0)$ is written as 
\begin{align}
B(s_0,\alpha_0) & = \sum_{s_{1:h},\alpha_{1:h-1} }  e^{ \beta \sum_{t=0}^{h-1} r_t } p(\alpha_{0:h-1},  s_{1:h} | s_0)  \nonumber \\
& = e^{ \beta r_0 } \pi_0 (\alpha_0 | s_0)  \sum\limits_{s_1,\alpha_1} p(s_1|s_0,\alpha_0) \sum\limits_{s_{2:h},\alpha_{2:h-1}}  e^{ \beta \sum\limits_{t=1}^{h-1} r_t } p(\alpha_{1:h-1},  s_{2:h} | s_1)  \nonumber \\
& =  e^{ \beta r_0 } \pi_0 (\alpha_0 | s_0) \sum\limits_{s_1} p(s_1|s_0,\alpha_0) \sum\limits_{\alpha_1} B(s_1,\alpha_1) 
\end{align}
More generally, we have the recursion 
\begin{align}
\label{eq:Bellman1}
B(s_t,\alpha_t) & =  e^{ \beta r_t } \pi_0 (\alpha_t | s_t)   \left( \sum\limits_{s_{t+1}}  p(s_{t+1}|s_t,\alpha_t)  \sum\limits_{\alpha_{s+1}}  B(s_{s+1},\alpha_{s+1}) \right) \\
& = e^{ \beta r_t } \pi_0 (\alpha_t | s_t)  \sum\limits_{s_{t+1}}  p(s_{t+1}|s_t,\alpha_t) A(s_{s+1})  
\label{eq:Betarecursion}
\end{align}
where for any terminal state $s \in \mathcal{T}$ (for which we take no further actions) 
$B(s,\alpha)$ is such that $\sum_{\alpha} B(s,\alpha) = 1$, which is consistent with the recursion.  


We now prove Proposition 2. 
From \eqref{eq:Bellman1} by taking logarithms of both sides, dividing by $\beta$ and by 
using the fact that $p(s'|s,\alpha) = \delta(s' - d(s,\alpha))$ we have that 
$$
\frac{1}{\beta}  \log B(s,\alpha) = r + \frac{1}{\beta} \log \pi_0 (\alpha | s)
+ \frac{1}{\beta} \log \sum\limits_{\alpha}  B( d(s,\alpha),\alpha)  
$$
Now set $M = \max_{\alpha}\{ \log  B( d(s,\alpha),\alpha) \}$ and let $\alpha_*$ be the 
action for we which this maximum is attained. The above is written as
as 
$$
\frac{1}{\beta}  \log B(s,\alpha) = r + \frac{1}{\beta} \log \pi_0 (\alpha | s) + 
\max_{\alpha}\{ \frac{1}{\beta} \log  B( d(s,\alpha),\alpha) \}
+ \frac{1}{\beta} \log ( 1 + \sum\limits_{\alpha \neq \alpha_*} e^{\log  B( d(s,\alpha),\alpha) - M } )  
$$
It holds that $0 \leq \log(1 + \sum_{\alpha \neq \alpha_*} e^{\log  B( d(s,\alpha),\alpha) - M }) \leq \log K$. Thus,  
by taking the limit $\beta \rightarrow \infty$ the terms $\frac{1}{\beta} \log \pi_0 (\alpha | s)$  and 
$\frac{1}{\beta} \log ( 1 + \sum_{\alpha \neq \alpha_*} e^{\log  B( d(s,\alpha),\alpha) - M } )$ 
tend to zero from which we conclude that $\frac{1}{\beta}  \log B(s,\alpha) \rightarrow Q_*(s,\alpha)$.

\bibliography{refs}
\bibliographystyle{apalike}

\end{document}